# Driver Drowsiness Detection System – An Approach By Machine Learning Application

Jagbeer Singh, Ritika Kanojia, Rishika Singh, Rishita Bansal, Sakshi Bansal,

Computer Science and Engineering Dept, Meerut Institute of Engineering and Technology, Meerut, India
ritika.kanojia.cs.2019@miet.ac.in, rishika.singh.cs.2019@miet.ac.in, rishita.bansal.cs.2019@miet.ac.in, sakshi.bansal.cs.2019@miet.ac.in
DOI: 10.47750/pnr.2022.13.S10.361

## Abstract

The majority of human deaths and injuries are caused by traffic accidents. A million people worldwide die each year due to traffic accident injuries, consistent with the World Health Organization. Drivers who do not receive enough sleep, rest, or who feel weary may fall asleep behind the wheel, endangering both themselves and other road users.

The research on road accidents specified that major road accidents occur due to drowsiness while driving. These days, it is observed that tired driving is the main reason to occur drowsiness. Now, drowsiness becomes the main principle for to increase in the number of road accidents. This becomes a major issue in a world which is very important to resolve as soon as possible. The predominant goal of all devices is to improve the performance to detect drowsiness in real time. Many devices were developed to detect drowsiness, which depend on different artificial intelligence algorithms. So, our research is also related to driver drowsiness detection which can identify the drowsiness of a driver by identifying the face and then followed by eye tracking. The extracted eye image is matched with the dataset by the system. With the help of the dataset, the system detected that if eyes were close for a certain range, it could ring an alarm to alert the driver and if the eyes were open after the alert, then it could continue tracking. If the eyes were open then the score that we set decreased and if the eyes were closed then the score increased. This paper focus to resolve the problem of drowsiness detection with an accuracy of 80% and helps to reduce road accidents.

## I. INTRODUCTION

Drowsiness refers to sleepiness. The drowsiness might also additionally ultimate for a couple of minutes however its consequences can be disastrous. The main cause of sleepiness is typically exhaustion, which reduces alertness and attention, although other causes include lack of concentration, medications, sleep issues, drinking alcohol, or shift work. They are unable to predict when sleep may strike. Even though falling asleep while driving is dangerous, being tired makes it difficult to drive safely even when you are awake. One in twenty drivers is said to have fallen asleep behind the wheel.

The most at risk for tired driving is truck and bus drivers with commutes of 10 to 12 hours. These people endanger other drivers more than they endanger themselves. Driving a long distance while sleep deprived might make you drowsy, as can driving when you need to sleep. In these scenarios, the driver's drowsiness has developed and is to blame for any accidents that occur on road.

According to National Highway Traffic Safety Administration (NHTSA), the police and hospital reports identified that 100,000 car accidents and over 1,500 deaths were caused due to drowsiness of drivers each year. Drowsy driving is thought to be responsible for approximately 1,550 fatalities, 71,000 injuries, and $12.5 billion in financial losses [4]. A sleepy driver was a factor in 697 fatalities in 2019. NHTSA acknowledges that it is challenging to quantify the precise number of accidents, or fatalities caused by drowsy driving and that the reported figures are underestimates [5].

Fortunately, it is now possible to recognise tiredness in a motorist and warn them before a collision. Drivers who are drowsy exhibit a variety of symptoms, such as frequent yawning, closed eyes for an extended period of time, and erratic lane changes. In recent years, methods for identifying driver drowsiness (DDD) have been effectively researched.

To prevent accidents, researchers have suggested a variety of ways to identify tiredness as soon as feasible. In our effort, detecting drowsiness begins with the identification of a face, followed by the identification of an eye's position and pattern of blinking. A "Shape predictor including 68 landmarks" is used to analyse faces. We utilise a camera—most likely a webcam in this instance—positioned in the direction of the driver's face to identify the driver's face and his or her facial landmarks in order to estimate the position of the driver's eye. To do this, it must use in-house image processing to examine every face and set of eyes. When the system locates the position of the eyes, it next determines whether they are open or closed and the blinking rate, which is how quickly the eyes are closing and opening. The alarm will sound, warning the driver, after a predetermined amount of time with the eyes closed. We begin with a score of zero for the eye open or closed; if the eye is closed, the score will rise, and if it is open, the score will fall. If the score exceeds a limit, then the alarm will ring to alert driver.

The remaining sections of this paper are structured as follows: Literature review, methodology, experimental result discussion, conclusion, and references.

## Literature Review

Several strategies were used in a tender to improve the efficiency and speed of the sleepiness detection procedure. The methods and strategies used in the past to identify drowsiness are the main topic of this section. The first method is based on driving patterns, which also take into account vehicle features, road conditions, and driving techniques. Calculating steering wheel movement or deviation from lane position will help you determine your driving style [6][7]. Driving requires constant control of the steering wheel to keep a car in its lane. Based on the correlation between tiredness and micro-adjustments, Krajewski et al.'s [6] detection of driver drowsiness had an accuracy of 86%. Additionally, it is possible to determine the driver's tiredness using a lane deviation approach. In this instance, the car's position in relation to a lane is tracked and examined to look for signs of sleepiness [8]. However, the methods based on driving patterns depend on the nature of the vehicle, the driver, and the road circumstances.

The alternative category of methods makes use of physiological detector data, such as electrocardiogram (ECG), electroencephalogram (EEG), and electrooculography (EOG) data. Information about the brain's activity is provided via EEG signals. Delta, theta, and nascence signals are three main signals used to gauge driver tiredness. When a driver is sleepy, theta and delta signals increase, while nascence signals barely change. This fashion is the most precise system, according to Mardi et al [9], with a delicacy rate of over 90. However, this system's biggest drawback is that it is obtrusive. The driver must have multiple detectors linked to them, which could be uncomfortable. Non-intrusive bio signal styles, on the other hand, are substantially less accurate.

The last possible solution is detecting facial features including yawning, face position, and eye blinking [11]. In the eye closure method, the condition of the driver is measured by counting the eye blink of the driver. The normal average duration of eye blink is 0.1s to 0.4s. It means that the eye will blink at least 2 or 3 times in one second. This is observed for a few seconds. When the driver is fatigued, the count will be less compared to normal conditions. So, we can detect whether the driver is fatigued or not. In our project, the camera is placed in front of face which helps to detect proper face position and eye blinking. Initially, the face is detected, and then eye, the closure process is recorded with the help of an open cv which detects the 68 landmarks of the face [10]. It is possible to tell if someone's eyes are open or closed using the Euclidean eye aspect ratio.

$$EAR = \frac{\|a_2-a_6\| + \|a_3-a_5\|}{2\|a_1-a_4\|}$$

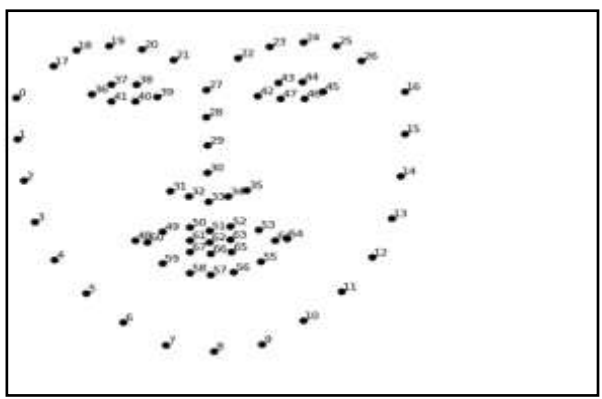

Fig.1. face landmarks detected by open cv

The system will next determine whether or not the eyes were open after detecting the eye. The alert would sound until the eyes were opened if the eyes were closed because the score would check to see whether it exceeded the predetermined score. As long as driver's eyes are open, the system will continue to track them.

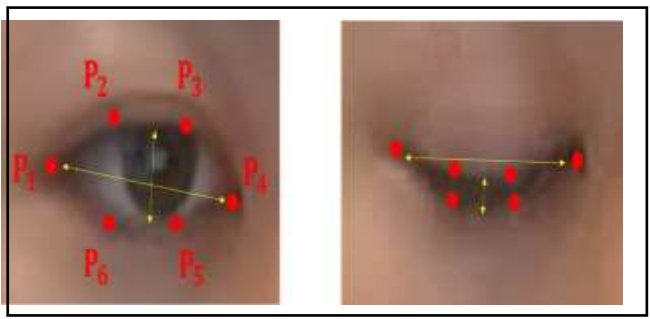

Fig.2. Six ocular landmarks are present both before and after the eyelids are closed.

Some methodological issues will occur In this paper like sample size estimation, classification of data, and detection of the eyelids. Future research will also be done In this paper, like detecting the yawning process and eye detection, which becomes more useful to detect the driver's drowsiness [12].

## II. METHODOLOGY

If we explain the general architecture of the model you will observe that it is very easy to operate this model because here only we have to capture the video of the face of the driver in the camera so that it will measure the scoring of blinking of the eyes and beep the alarm accordingly [13][14][15].

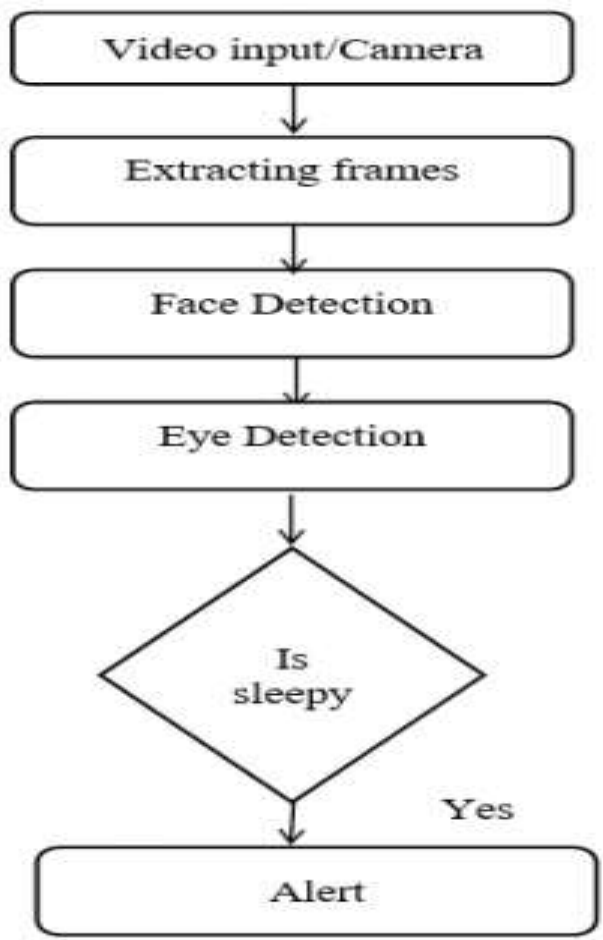

Invasive machine vision-based concepts are used in the development of the Driver Drowsiness system. In this, the driver's face is the focal point of a webcam that can be used to identify his face. After identifying the face, it concentrates on the eyes and their state, such as whether they are open or closed. The eyes are moved to look for signs of weariness [36] [37]. Additionally, the driver receives a warning signal if weariness is found so that he can make necessary adjustments[16[17][18].

For the purpose of sleepiness detection In this paper, Python is used. The system only deals with the face as a specific bodily part. A webcam is positioned in front of the driver's face to record the input video. The algorithm will assume that drivers are sleeping if a face is not found after several frames. With the use of 68 facial landmarks, OpenCV is utilized to identify face and eye. If the eye is open or close, it can be determined using the Euclidean eye aspect ratio [38] [39]. The system will examine driver's face and eyes. If the eye is open or closed will then be determined. The alarm will sound to alert the driver if the specified time interval is shorter than the time interval during which the eyes are closed. In the event that driver's eyes are opened, the device will continue to track their eyes [19][20][21].

We also employ PERCLOS, which stands for "the percentage of eyelid closure over the pupil over time" and represents gradual eyelid closures as opposed to blinks. The system as a whole is utilized to gauge Perclos, and depending on the scoring of Perclos, the beep begins to alarm [33], [34] [35].
In this paper, we import some library like as mentioned below:-

CV2: The open-source library known as OpenCV is employed in the fields of machine learning, computer vision, and image processing. It can analyze pictures and movies to spot people, objects, etc.[22][23][24].
OS: Python's OS module is used to communicate with operating system functions. It falls under one of Python's common utility modules. This module gives users a portable means to access the operating system-specific features [25][26][27].

Keras: It is a python-based, high-level neural network library that can function with tensor flow. This module focuses on user experience and is highly adopted by the industry, seamlessly running on CPU as well as GPU.[28][29][30].
NumPy: To work with arrays, one uses a Python library. It includes matrices and linear algebra functions for use.

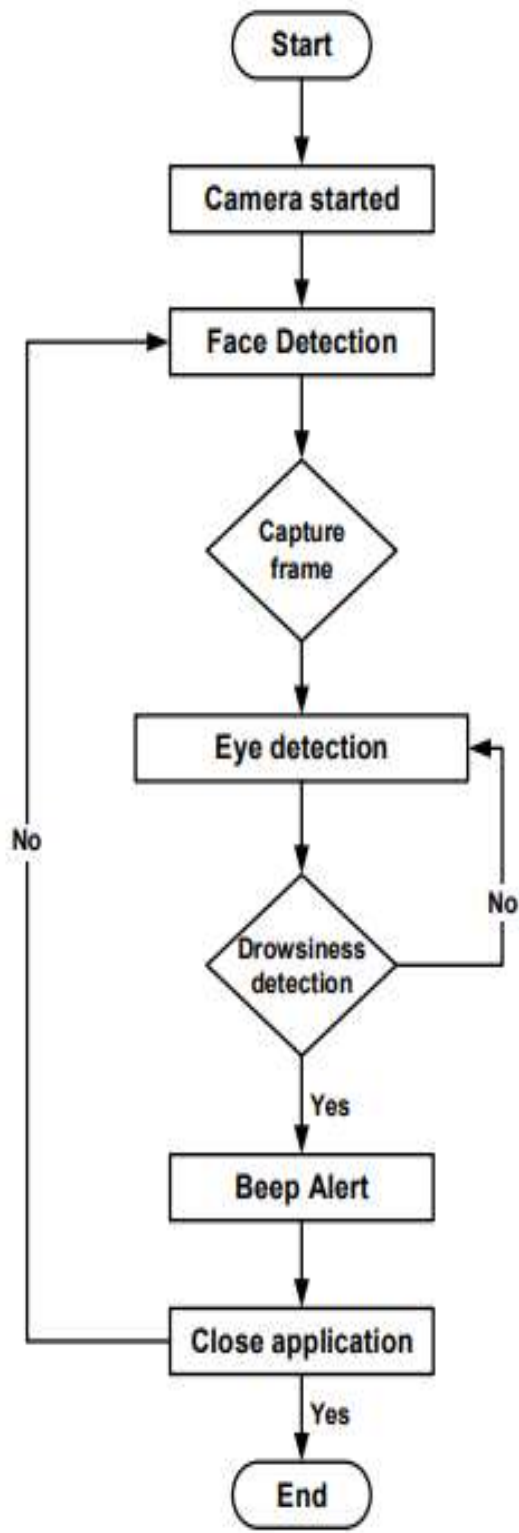

Pygame: It is a Python cross-platform collection. Video games are created using it. It consists of music and graphic design libraries for the Python programming language.[31][32].

Matplotlib: It is a low-level graph plotting library in python language. It represents a visualizing graph.[33][34].

### III.  OUTCOME SCREENS SHOT

1.  On running the program this first screen is shown

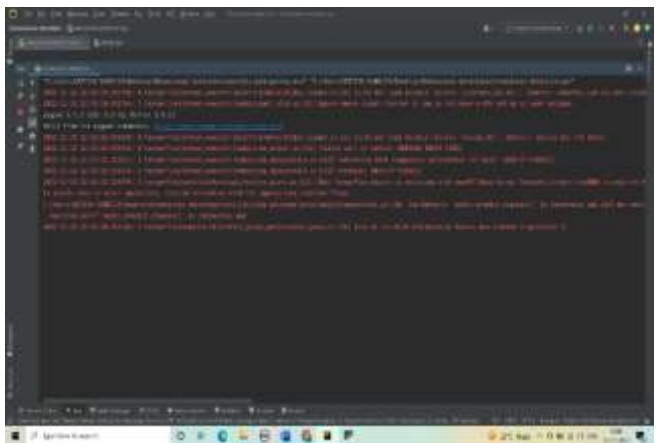

2. Now camera is open and detect the face, the eyes are open so score is zero by using with that things and this is an example of without wear glasses

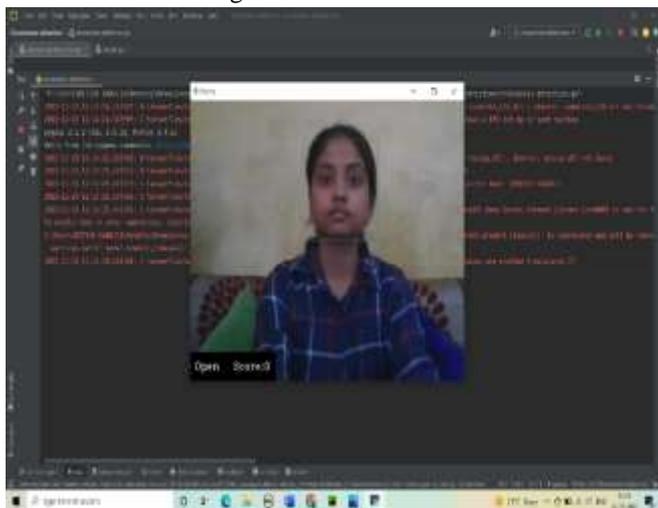

3. The eyes were closed so score will increase

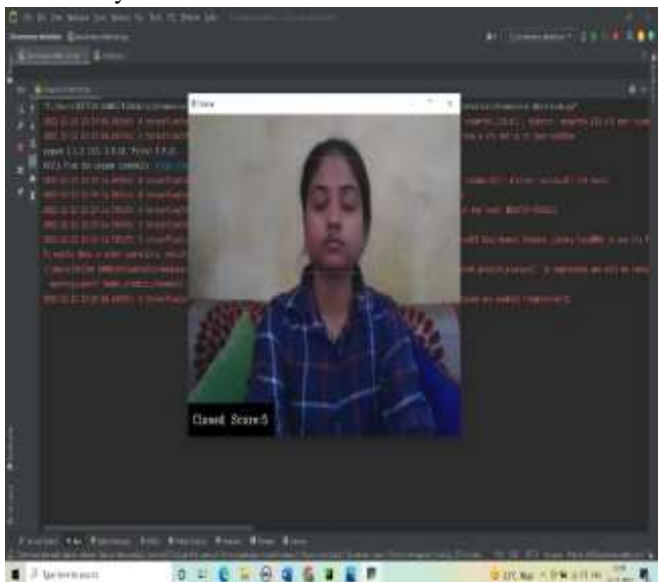

4. The eyes were closed and score is greater than the set score which is '10' so the alarm will beep to alert the driver

Journal of Pharmaceutical Negative Results ¦ Volume 13 ¦ Special Issue 10 ¦ 2022    3007

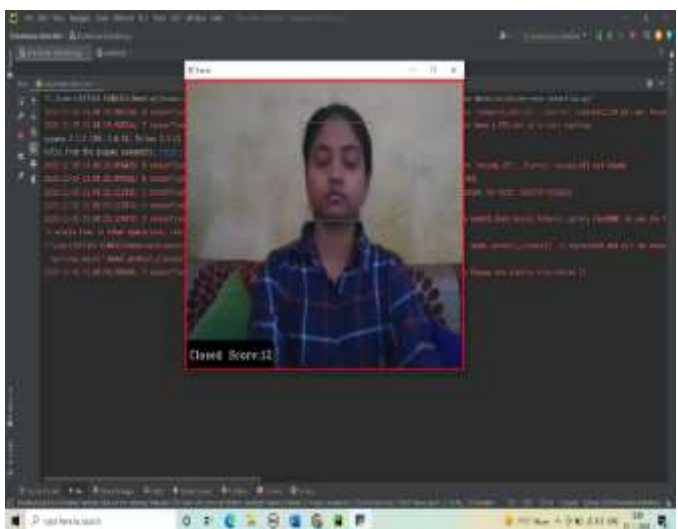
5. Now, the eyes had been open so the rating will lower and forestall the alarm

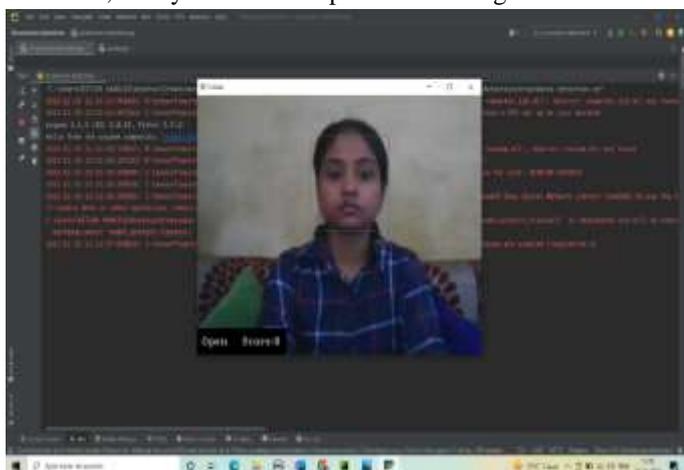
6. Now, the example of detecting the eyes with wear glasses and the eyes were open so score is '0'

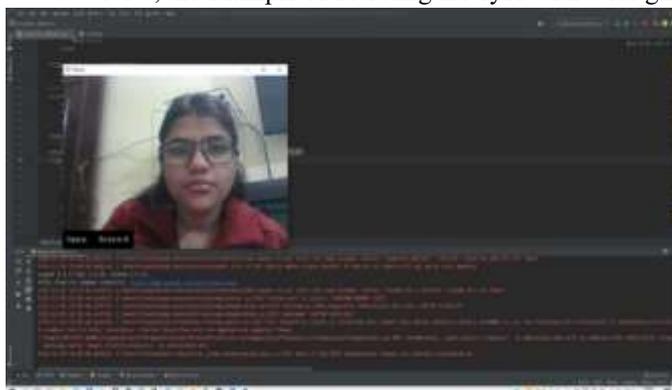
7. The eyes were close so score will increase but it not reaches the set score value so the alarm will not beep

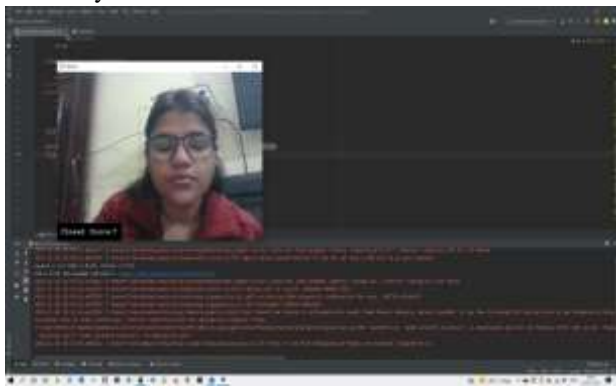

8. The score is '14' which is greater than the set score value so the alarm will beep [38][39].

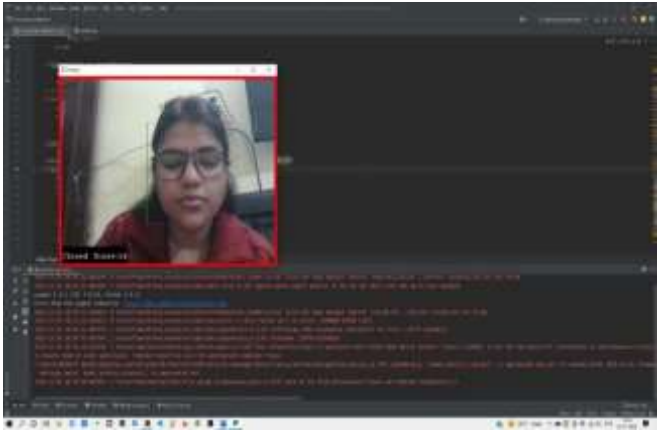

9. On keyboard interrupt the camera will close and the program will stop [35][36][37].

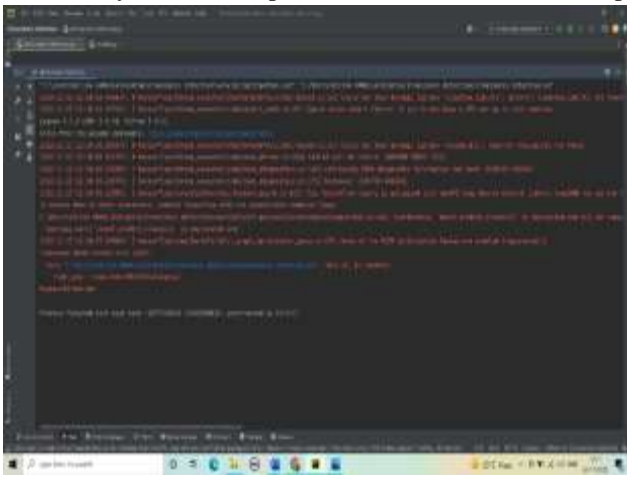

## IV. RESULT ANALYSIS

The main approach to detecting any image features extraction from facial landmarks. Facial landmarks are commonly known as the subset of the shape predictor problem as this can be used to localize any area of interest like the eye, nose, and mouth along with the shape of the subject. The dlib library consists of a facial-landmark detector that's used to discover 68(a, b) coordinates.

- Open Eye Coordinates

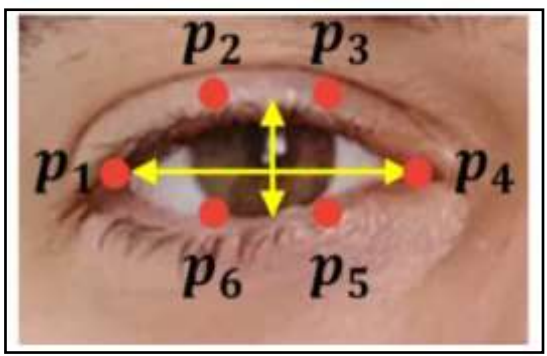

Fig.3. In this eye coordinates a1, a2, a3, a4, a5, and a6 used for measure eye aspect ratio (EAR) for an open eye is approx. 0.24.

- Close Eye Coordinates

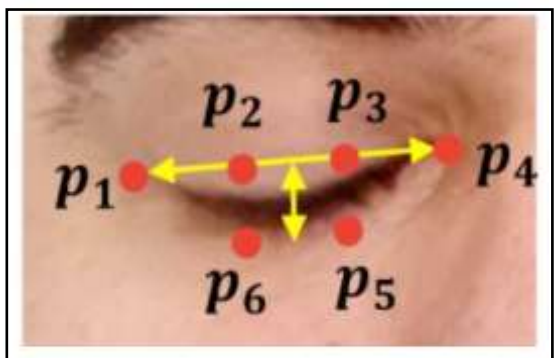

Fig.4. Eye aspect ratio (EAR) for a close eye is approx. 0.15.

| INDIVIDUAL | EAR THRESHOLD | ALARM SENSITIVITY | LIGHT | REMARKS | DROWSINESS DETECTION ALARM |
|---|---|---|---|---|---|
| A | 0.2 | 48 | Bright | Normal | 3 out of 3 |
| A | 0.2 | 48 | Dim | Normal | 3 out of 3 |
| A | 0.2 | 48 | Bright | Wear sunglasses | 0 out of 3 |
| B | 0.25 | 43 | Bright | Normal | 3 out of 3 |
| B | 0.25 | 43 | Dim | Normal | 3 out of 3 |
| B | 0.25 | 43 | Dim | Rainy weather | 2 out of 3 |
| C | 0.22 | 48 | Bright | Wear glasses | 3 out of 3 |
| C | 0.22 | 48 | Dim | Wear glasses | 3 out of 3 |
| C | 0.22 | 48 | Very Dim | Night drive | 1 out of 3 |
| C | 0.22 | 48 | Very Dim | Normal | 3 out of 3 |

The whole test was conducted 10 times with different parameters such as surrounding light, different drivers, and alarm sensitivity. The below table shows the testing parameters for the accuracy test. The tests were conducted to observe the accuracy of the whole project by using the accuracy formula:

$$CR = (C/A) \times 100\%$$

Where CR is the correct rate, C is the number of correct tests and A is the number of tests. 8 out of 10 tests were running successfully and smoothly while 2 tests failed due to bad light conditions at night. Hence, the resultant accuracy of the project is around 80%. The light conditions during the experiment affected the resultant accuracy and the output. The main factor is the brightness of the light,

Test parameters for accuracy test

which influences the output of the drowsiness detection system Therefore, our project's accuracy is 80% on average.

V. CONCLUSION

In conclusion, driver drowsiness detection system technology is a vehicle protection technology to assist to save your injuries resulting from drowsy drivers. It is important to detect and alert the driver early before any unwanted accidents happen that may lead to death. The proposed system can detect driver drowsiness levels using an image processing technique that calculates and measures Eye Aspect Ratio, in other words, size of driver's eye. The data on Eye Aspect Ratio has to be gathered to determine the threshold value that indicates whenever a driver is experiencing drowsiness. An alert system using an alarm is vital as it helps in reducing the variety of injuries because of drowsy using subsequently decreasing the overall of a car that crashed annually.

As for now, the detection system can come across the drowsiness of the equal driving force over and over with very minimum limitation. The alarm is also running properly and might cause a valid alarm to alert the driver. However, threshold frames to trigger the alarm may vary due to different Eye Aspect Ratios (EAR) in every person. Several recommendations are suggested for future works in this area. First, the system should be able to automatically determine the threshold of eye aspect ratio when a person is experiencing drowsiness without setting it first for each separate individual after several testing.

This is because some people tend to have higher precautions and awareness towards road safety thus wanting a more sensitive and frequent alarm alert system.